# Characterizing Lidar Point-Cloud Adversities Using a Vector Field Visualization

Daniel Choate and Jason H. Rife, *Tufts University*

**BIOGRAPHY**

**Daniel Choate** is a student in the Mechanical Engineering Ph.D. program at Tufts University in Medford, MA. He works in the Automated Systems and Robotics Laboratory (ASAR) with Dr. Jason Rife. He received his B.S. degree in Mechanical Engineering from Union College in 2023.

**Jason Rife** is a Professor and Chair of the Department of Mechanical Engineering at Tufts University in Medford, Massachusetts. He directs the Automated Systems and Robotics Laboratory (ASAR), which applies theory and experiment to characterize integrity of autonomous vehicle systems. He received his B.S in Mechanical and Aerospace Engineering from Cornell University and his M.S. and Ph.D. degrees in Mechanical Engineering from Stanford University.

**ABSTRACT**

In this paper we introduce a visualization methodology to aid a human analyst in classifying adversity modes that impact lidar scan matching. Our methodology is intended for offline rather than real-time analysis. The method generates a vector-field plot that characterizes local discrepancies between a pair of registered point clouds. The vector field plot reveals patterns that would be difficult for the analyst to extract from raw point-cloud data. After introducing our methodology, we apply the process to two proof-of-concept examples: one a simulation study and the other a field experiment. For both data sets, a human analyst was able to reason about a series of adversity mechanisms and iteratively remove those mechanisms from the raw data, to help focus attention on progressively smaller discrepancies.

## 1    INTRODUCTION

This paper introduces a visualization technique for offline identification and characterization of lidar adversities that degrade scan matching. Our approach works iteratively, calculating a discrepancy-vector field from a pair of point clouds to aid a human observer in proposing a discrepancy-mitigation model, which might account for the patterns observed in the vector-field plot.

Our work is motivated by the applications of lidar to automated driving. Lidar sensors have been used in many commercial and prototype automated vehicles as a component within a larger sensing suite, often including GNSS, INS, wheel odometry, and sometimes also cameras or radar (Ilci & Toth, 2020). Lidar sensors are advantageous in that they can create a high-resolution 3D representation of the surroundings (Olson, 2009); this 3D representation of the scene can be used for localization (Hassani & Joerger, 2021), simultaneous localization and mapping (SLAM) (Cadena *et al.*, 2016), and for collision avoidance (Wei *et al.*, 2018). In this paper, we will focus on using lidar for localization, by registration of one lidar image with another (relative localization) or with a high-definition map (absolute localization).

When lidar registration infers vehicle pose, meaning its rotation and translation, random noise and systematic biases propagate into errors in the estimated vehicle pose. To date, several error mechanisms have been identified for lidar registration (Rife & McDermott, 2024) as well as change detection in 3D scanning (Salimpour *et. al.,* 2022; Nurunnabi *et. al.,* 2015). There has also been previous work regarding mitigation of various error mechanisms related to scene geometry (Li & Wang, 2020), continuous change of a scene such as moveable objects (Jeong & Kim 2023), and historical terrain anomalies (Storch *et. al.,* 2023). Adverse weather and off-road environments have also been shown to degrade lidar scan-matching (Fu & Yu, 2020; Nagai *et. al.,* 2024).

Identifying additional error mechanisms for lidar scan registration, or *scan matching*, remains an open research topic. To aid in identifying various adversities in lidar data that lead to registration errors, we describe in this paper a new process for human-in-the-loop characterization.

In this paper, we assert that each adversity mechanism will broadly impact all scan matching algorithms, even though specific error levels may vary from one scan-matching algorithm to another. For example, iterative closest point or ICP methods (Segal *et al.,* 2009), normal distributions transform or NDT (Biber & Straßer, 2003), iterative closest ellipsoidal transform or ICET (McDermott & Rife, 2023), and lidar odometry and mapping or LOAM (Shan & Englot, 2018) emphasize different aspects of lidar point clouds in order to perform scan matching, so they each obtain different results and different errors when estimating pose. Nonetheless, all these methods suffer errors due to the presence of moving objects, such as pedestrians or motor vehicles, unless the corresponding points are pruned from lidar data sets prior to scan matching.

With the idea that point-cloud adversities impact a wide-range of algorithms, we propose an offline, human-in-the-loop methodology to identify and characterize them. Our key contribution is to introduce discrepancy-vector fields to help the human analyst to visualize and reason about possible lidar adversities. Our implementation starts by aligning a pair of point clouds using ground truth (e.g., using carrier-phase differential GNSS for experimental data). Next, a voxel grid is introduced. For the points within each voxel, a mean location is computed for each of the two clouds. The voxel-based means are differenced to create a discrepancy vector. A vector field is then created by compiling the discrepancy vectors over all voxels. A human analyst observes this vector-field visualization to identify patterns in the field. The human observer then hypothesizes the mechanism causing the largest observed discrepancies and proposes a pruning rule (or other transformation) to mitigate the discrepancy. The hypothesized rule can then be tested to check whether it indeed reduces the magnitude of the discrepancy vectors in the region of interest. In short, this methodology provides a systematic though labor-intensive offline characterization of discrepancies between a pair of point clouds.

The main goal of the method is to troubleshoot and identify impactful discrepancies within a specific scene. A secondary goal of the method is to visualize representative adversities in lidar point clouds, to aid in their explanation. In prior work, it has been common to hypothesize error modes and mitigate them in the context of a specific scan-matching algorithm, with validation conducted by statistical analysis on the improvement in the pose estimate (Rife & McDermott, 2024; Joerger *et al.,* 2022). Our approach can provide a statistical description, but also an interpretable visualization, that is intended to be somewhat algorithm agnostic.

In the next section of the paper, we provide more detail on the proposed methodology for visualizing adversities in lidar data. Subsequently, we apply the method to two types of data: (i) simulated data where exact truth is known and (ii) experimental data where ground truth must be inferred. A brief discussion and summary conclude the paper.

## 2   METHODOLOGY

In this paper we introduce a visualization approach to aid in identifying and reasoning about sources of lidar adversity. The basic idea of the visualization is to align a pair of point clouds (or a point cloud and a high-definition map) and then look for local discrepancies between the clouds. The details of our approach are illustrated as a block diagram in Figure 1.

In the diagram, the key feature is the human inspection step, in which an analyst examines a vector-field plot to localize and characterize patterns of discrepancies in the lidar data. The human analyst proposes adversity models with the goal of accounting for the patterns observed in the vector-field plot. As shown in the block diagram, hypotheses are iteratively evaluated, with the human analyst rejecting invalid hypotheses and preserving valid ones. The key data that mediates decision making on each iteration is the discrepancy-vector field.

In order to understand the discrepancy-vector visualization, it makes sense to start with the input data, which consist of a pair of point clouds, labeled as Pt. Cloud 1 and Pt. Cloud 2 in Figure 1. A point cloud represents an image composed of thousands of lidar returns, with each point $i$ described by a three-dimensional vector $\{x_i, y_i, z_i\}$. The first point cloud represents a current snapshot of a scene; the second point cloud represents a past snapshot of the scene, either taken by the same vehicle (e.g., to support dead reckoning) or drawn from a high-definition map (e.g., to support absolute positioning).

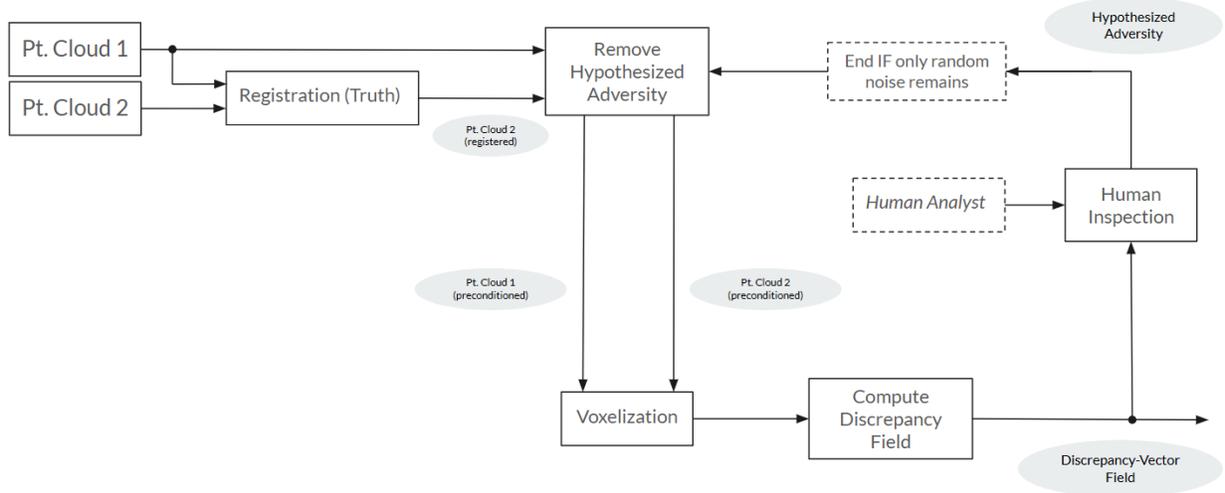

**FIGURE 1**
*Flow diagram displaying the process of vector-field visualizations. The methodology takes two LIDAR point clouds as inputs and produces a discrepancy-vector field for visualization by a human analyst.*

If the point clouds capture roughly the same scene, then they can be aligned so that static features appear at the same coordinates in both point clouds. This alignment, or *registration*, process requires determination of pose, meaning the relative orientation and translational offset between the two point-clouds. By default, each point cloud has its origin at the center of the lidar unit, and its basis vectors attached to the lidar unit, such that the coordinate system changes whenever the lidar undergoes rigid body translation and/or rotation.

The registration process ideally uses truth (for simulated data sets) or ground truth (for experimental data) to obtain the alignment. For instance, in an experiment where GPS and inertial data are available, the second image can be transformed to align with the first image using relative position data from the GPS and relative orientation data from the inertial navigation system. In the event that ground truth is not available, registration can also be conducted using the lidar data only, with a scan-matching algorithm aligning image features as well as possible, for instance through algorithms like NDT (Biber & Straßer, 2003), ICP (Segal *et al.*, 2009), or ICET (McDermott & Rife, 2023).

Once the second image is aligned to the first, any prior hypothesis-driven models can be applied to the data to mitigate discrepancies (via the "Remove Hypothesized Adversity" block in Figure 1). During the first iteration through the figure, the human analyst has not seen the data, so hypothesis-driven models don't yet exist, and this block just acts as a pass through.

As a next step, local discrepancies are computed through a voxelization process. The voxelization divides the entire scene into equal sized volumetric bins and places the lidar points from each scan within a specific bin. Although the volumetric bins could be cubes, defined in Cartesian Coordinates $\{X, Y, Z\}$, we use spherical volume elements, or *voxels*, defined in radius, azimuth, and elevation $\{r, \theta, \phi\}$. After lidar points are binned, each voxel contains one subset of returns from Point Cloud 1 and a second subset from Point Cloud 2. Ideally, with perfectly aligned scenes and no discrepancies, the two sets of returns appear to be identical and, by extension, the mean location of points within a given voxel should be the same for Point Cloud 1 and Point Cloud 2. Due to noise and other adversities, the voxel means are not identical in practice. The local level of misalignment within voxel $i$ can be computed as the difference between the mean $\{^1\bar{X}_i, {}^1\bar{Y}_i, {}^1\bar{Z}_i\}$ for Point Cloud 1 and the mean $\{^2\bar{X}_i, {}^2\bar{Y}_i, {}^2\bar{Z}_i\}$ for Point Cloud 2. Based on this definition, the components of the discrepancy vector are

$$\{U_i, V_i, W_i\} = \{^2\bar{X}_i, {}^2\bar{Y}_i, {}^2\bar{Z}_i\} - \{^1\bar{X}_i, {}^1\bar{Y}_i, {}^1\bar{Z}_i\} \tag{1}$$

Since a local discrepancy vector can be computed within each voxel, a vector field can be visualized by plotting the discrepancy $\{U_i, V_i, W_i\}$ over all voxels $i$.

A vector field is a concept from calculus (Anton, *et. al.*, 2020) that defines a vector value at each point in a continuous space. An example in physics is the velocity field of a fluid flow, such as for the airflow around a wing, where the air at each point in space has a 3D velocity vector that describes its local direction and speed of motion. In the case of a velocity field, the data is a continuum. In the case of our voxelized point clouds, equation (1) generates a field of discrepancy vectors, but because the space is discretized, the field of vectors is not a continuum. Rather, the discrepancy field is discrete, with vectors defined at the centroid of the points from cloud 1 within that voxel. In this sense, the tail of each visualized vector represents the centroid from point cloud 1, and the tip of the vector represents the centroid of point cloud 2 within that voxel.

As a means of conceptualizing a discrepancy-vector field, consider three example cases shown in Figure 2. Each case shows a hypothetical 3x3 grid of nine voxels total. The first example case (left) features uniform discrepancy vectors across the grid; this special case represents a bad registration, where the discrepancies can be fixed by shifting the point clouds, moving the second cloud opposite the direction of the discrepancy vectors to reduce their length. The second example case (middle) is one in which large opposing vectors are shown along the top and bottom rows. This is a case of a discrepancy that cannot be explained by bad registration, since there is no translation or rotation that will reduce the magnitude of the discrepancies. Instead, a human analyst might identify this as a systematic error, perhaps a calibration bias that introduces an incorrect scaling into one point cloud but not the other. If the human analyst can successfully reason about the cause of discrepancies, then the human analyst can mitigate the large biases such that only small residual vectors remain. This is the case in the third example (right side of Figure 2), where discrepancy vectors have small magnitudes and point in seemingly random directions.

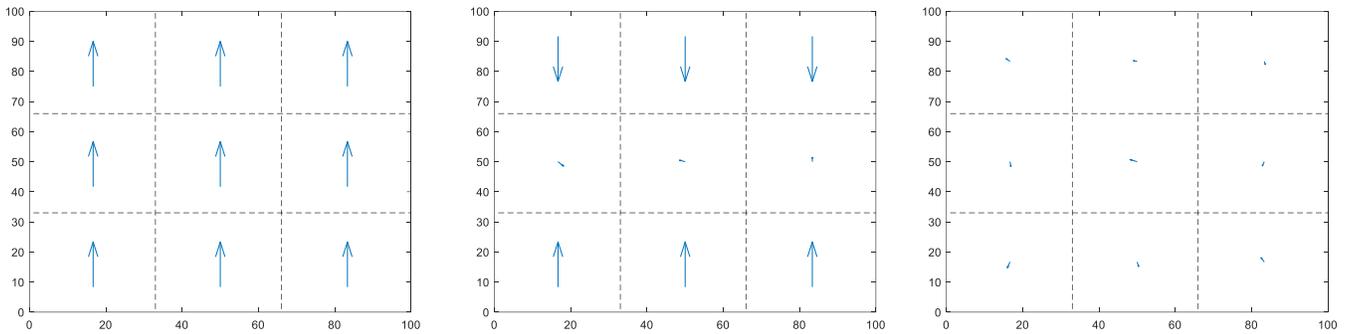

**FIGURE 2**
*Three examples of a 2D vector field, defined over a three-by-three voxel grid. (Left) Idealized discrepancy vectors for pure translation, which can be corrected by better registration; (Middle) Discrepancy vectors that cannot be corrected by registration, such that the human analyst will generate a hypothesis about the physical basis for the discrepancy; (Right) A random vector field of small magnitude – the ideal outcome after removal of the hypothesized condition.*

In the visualization-driven methodology illustrated in Figure 1, the human analyst is the workhorse that converts the discrepancy-vector field into a hypothesis about the physical basis for the discrepancies. That hypothesis can be used to generate a mitigation method for the proposed physical mechanism, for instance, a method for detecting the issue and excluding data from corrupted voxels. The mitigation method (or rather a series of all validated mitigations) is applied to the raw data via the "Remove Hypothesized Adversity" block in Figure 1.

After mitigating adversities, the discrepancy-vector field is recomputed and shown to the human analyst. Three things may happen at this point in the process. First, the magnitude of the discrepancy field may be somewhat reduced, but with salient patterns remaining. In this case, the proposed mitigation method was effective, but the scene still exhibits evidence of other adversities. Second, the magnitude and patterns of the discrepancy field may be unchanged, in which case the analyst deems the proposed mitigation ineffective. Finally, it is possible that the vector field contains only small-magnitude discrepancy vectors oriented randomly, as shown on the right side of Figure 2. In this case new patterns are not discernable, so the iteration process stops.

## 3  APPLICATIONS

The effectiveness of a vector-field visualization was explored through application to two lidar examples: one simulated and one experimental. The simulated scene allows for an analysis in which the ground truth is exact and the registration therefore ideal; the experimental scene allows for an analysis where the data includes real-world noise and distortions that do not appear in simulation. This section explains the two examples, their discrepancy-vector fields, and the process of identifying lidar-imaging adversities that introduce discrepancies between point clouds pairs formed at slightly different locations or at different times.

### 3.1 Generation of a Simulated Scene

As a first means of evaluating our vector-field methodology, a simulated scene was created using engineering drawing software (SOLIDWORKS). As shown in Figure 3, the simulated scene roughly models a road intersection featuring four "buildings," three modeled as rectangular prisms and one as a cylinder. The four structures are ten meters in height with rectangular footprints of 5x5 m², 8x5 m², 7x5 m² for the prisms and with a 5 m diameter for the cylinder. The dimensions of the ground plane are set arbitrarily to 26x26 m². Synthetic lidar data were generated for this scene using MATLAB. The lidar sensor was simulated in the middle of the scene, shown by the red star marker in Figure 3. The lidar is positioned three meters above the ground (at the roof height of a large truck or van). The MATLAB sensor model featured 80 channels separated by 0.4 degrees (spanning elevations between -22° and 10°). The resulting point cloud is shown in Figure 4. The point cloud captures the ground and part of the buildings (but not their roofs) because of the simulated lidar's field-of-view constraints in the elevation direction. To emphasize adversities due to geometric structures in the scene, the MATLAB simulation was configured to suppress random measurement noise.

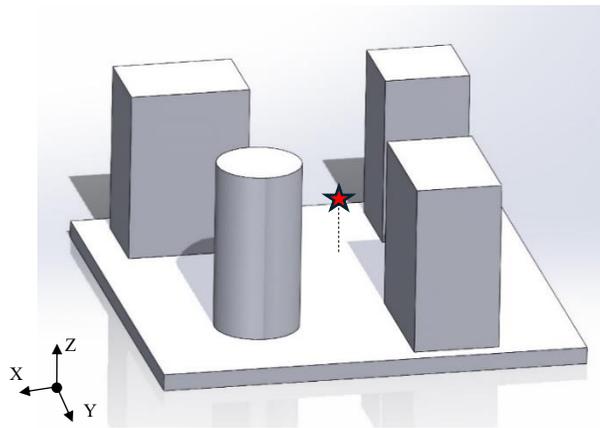

**FIGURE 3**
*An image of the SOLIDWORKS simulated scene, marking the placement of a theoretical lidar sensor.*

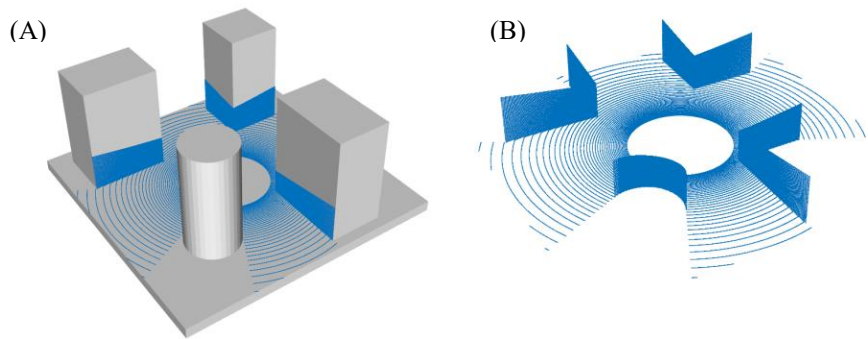

**FIGURE 4**
*Simulated point cloud generated at initial lidar location, shown (A) superimposed on a rendering of the scene and (B) absent the rendered scene.*

For the purpose of visualizing discrepancies, we generated a pair of point clouds from the scene to represent motion of the lidar for a vehicle rounding a curve. The initial position (red star in Figure 3) was $\{x_1, y_1, z_1\} = \{0,0,3\}\ m$ and the second position was $\{x_2, y_2, z_2\} = \{1,1,3\}\ m$ after a pitch-roll-yaw rotation of $\{\phi, \theta, \psi\} = \{0,0,0.05\}\ rad$. The MATLAB simulation tool assumed the entire point cloud was captured instantaneously, which is a reasonable assumption for low-speed motion, but which neglects the fact that today's rotating lidars typically collect lidar points using a rotating mechanism that completes a full rotation over a period of about 0.1 seconds. The finite rotation speed of the lidar rotor introduces motion distortion effects sometimes called *rolling shutter* (McDermott & Rife, 2023).

### 3.2 Application of Visualization Methodology to Simulated Data

The visualization methodology of Figure 1 was subsequently applied to the pair of point clouds generated from the simulated scene of Figure 3. Since the displacement and rotation of the lidar between images is known exactly, the "Registration" block can transform the second point cloud precisely to align with the first, such that static features in the scene are coincident. The aligned point clouds (red and blue) are shown in Figure 5. The alignment is evident in that the edges of the buildings are crisply defined for both point clouds. Note that the circular "hole" in the middle of the point clouds does not correspond to a feature in the environment; rather, this hole represents the elevation angle constraints of the lidar unit, which cannot sample the ground plane below an angle of -22°. The center of the circular hole corresponds to the lidar location, which was different when sampling each point cloud.

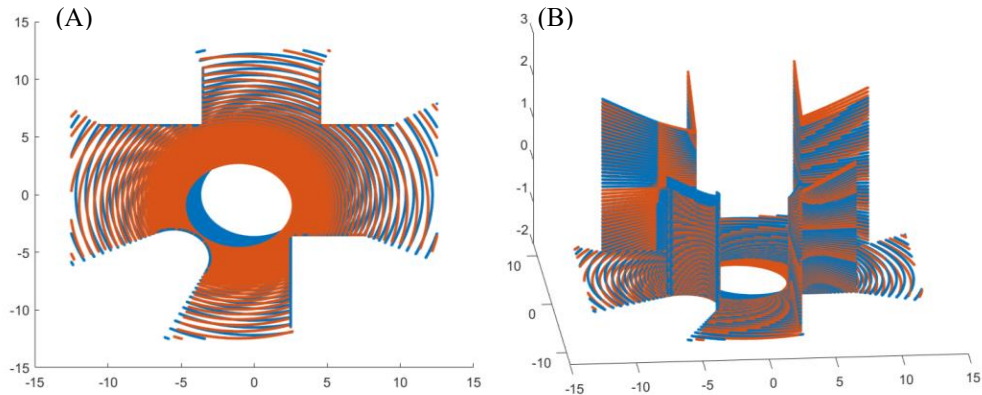

**FIGURE 5**

*Registration of two simulated point clouds (red and blue) shown (A) from above and (B) in perspective.*

The next major step of the methodology involves voxelization. (Note that the "Remove Hypothesized Adversity" block in Figure 1 acts as a pass-through initially, since the human analyst has not yet reasoned about the visualization or proposed any hypotheses to explain discrepancies.) In the voxelization process, we selected a spherical grid centered at the origin of the first point cloud. Each grid cell is a frustrum, meaning a semi-infinite wedge extending outward from the origin in the range direction, bounded within a side-to-side band of azimuth angles and a top-to-bottom band of elevation angles. Specifically, the full measurement space was divided into 36 azimuth bins (each of 10° in width, covering a full 360° azimuth range) and 9 elevation bins (each of 7.5° in height, covering a 37.5° elevation range, slightly larger than the sensor's 30° field-of-view cutoff in elevation). Based on empirical observation, discrepancy vectors are more easily characterized using a spherical voxel grid rather than a Cartesian grid.

After voxelization, the discrepancy-vector field is computed using (1). The resulting difference field is shown in Figure 6. A salient detail is that discrepancy vectors computed with (1) use cluster-mean locations framed in Cartesian coordinates, even though the cluster boundaries are defined using spherical voxels.

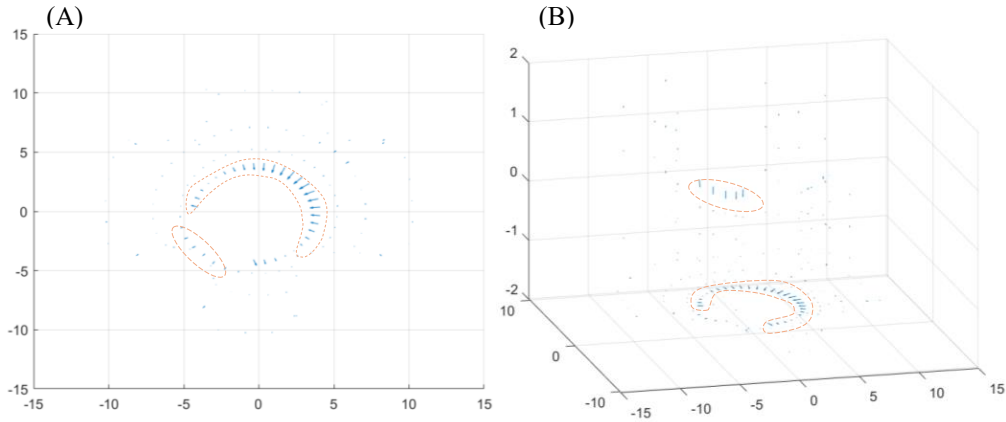

**FIGURE 6**
*Initial discrepancy vector field shown (A) from above and (B) in perspective. Discrepancies indicate differences between the mean position of red and blue points (see Figure 5) within each spherical voxel. Red dashed contours highlight the most prominent discrepancies considered by the analyst.*

Next, the human inspection step takes place. The human analyst identifies prominent patterns within the vector field. In the case of Figure 6, the human analyst marked two groups of discrepancy-vectors with a red dashed contour. One contour forming a ring-like pattern in Figure 6 appears to correspond to the field-of-view boundary for the sensor. Based on this observation, the analyst hypothesized that the discrepancy magnitude might be reduced by excluding points from one cloud that were outside the field of view of the other. The points excluded by this hypothesis are shown in red and blue in Figure 7, while the preserved points are shown in gray. The excluded points either fall on the ground plane (at the lower elevation limit) or high on buildings (at the upper elevation limit).

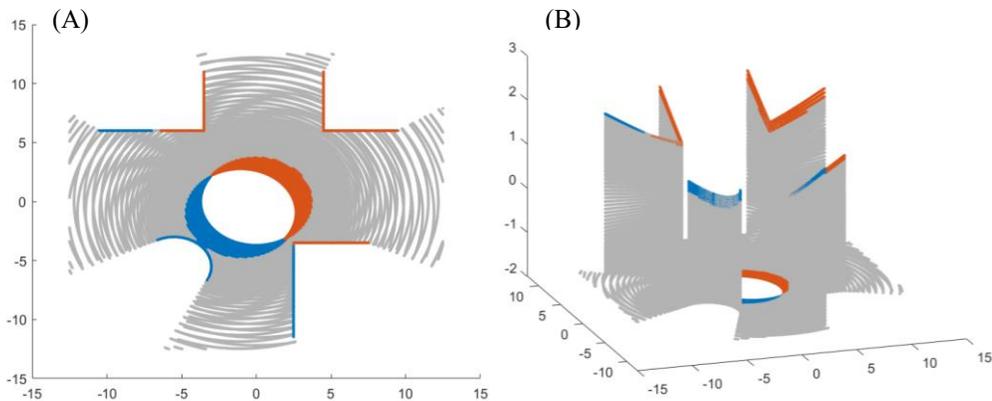

**FIGURE 7**
*Points in each cloud identified as lying outside the field-of-view associated with the other cloud, shown (A) from above and (B) in perspective.*

After implementing this exclusion rule, removing points from one cloud that are not within the field-of-view associated with the other, the pair of point clouds can be processed by the "Remove Hypothesized Adversity" block in Figure 1. Progressing around the processing loop, the human analyst is presented with a new discrepancy-vector field. The comparison of the original and new vector fields is illustrated in Figure 8. The large discrepancies visible in the initial vector field (left side of figure) are removed in the revised vector field (right side).

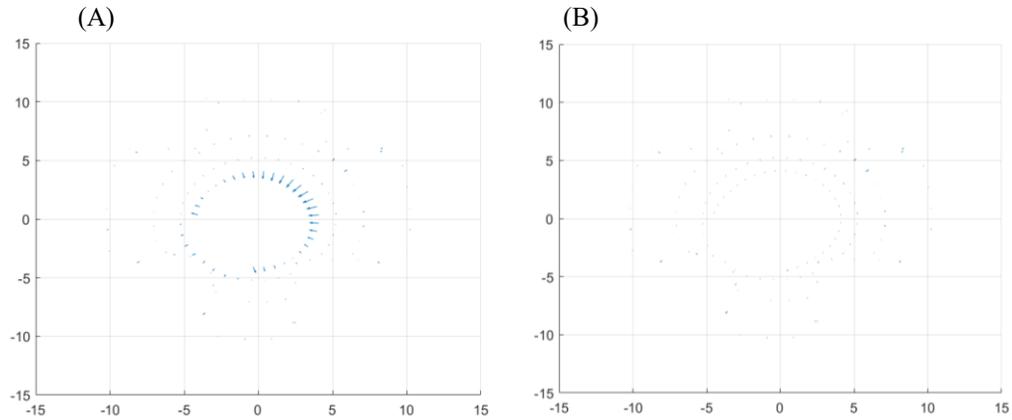

**FIGURE 8**
*Comparison of discrepancy-vector field (A) before and (B) after field-of-view mitigation.*

The question remains if other discrepancies can be identified by drilling down farther. To answer this question, the vectors in Figure 8B can be scaled. After magnification, patterns are again visible in the vector field, as shown in Figure 9A and B. The largest vectors seen in the magnified plot occur where the buildings meet the ground. To better visualize one of these regions, we identify the voxels generating the large arrows in the lower-left corner of Figure 9A. These voxels are shown as wireframes (dashed lines) in Figure 9C and D.

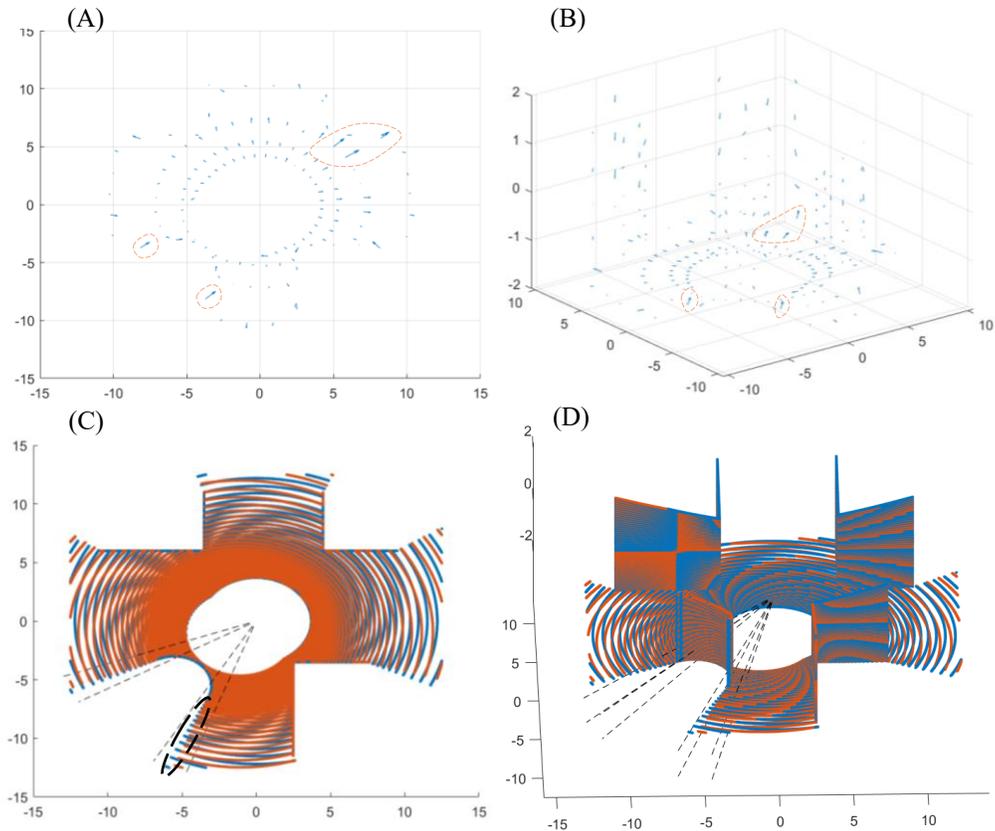

**FIGURE 9**
*Information used in second iteration, where the analyst identifies clusters as red dashed contours, shown (A) from above and (B) in perspective. The voxels containing the large discrepancies are shown as wireframes, with voxel boundaries superimposed on the registered point clouds, shown (C) from above and (D) in perspective.*

Looking at the voxels with the largest discrepancies (see Figure 9C and D) indicates a potential adversity associated with the edges of the shadow cast by the cylindrical structure. Shadowing errors, that occur where one object casts a shadow on another object or on the ground plane, are known to introduce spurious features (McDermott & Rife, 2022). Shadow errors can occur in lidar imaging of curved surfaces, when part of an object blocks another part of the same object from view (Rife & McDermott, 2024).

Previous work highlights solutions to target and mitigate occlusion and shadowing effects (McDermott & Rife, 2022; Shimojo *et. al.,* 1989). Focusing on the shadows cast by the cylinder, the analyst implemented the shadow-mitigation algorithm proposed by McDermott and Rife in the "Remove Hypothesized Adversity" block in Figure 1. The shadow-mitigation algorithm removes points from one point cloud if they might be blocked from sight from the lidar location for the second point cloud. The additional points removed in this step are shown in Figure 10, where removed points are shown in color and preserved points in gray. On the third loop through Figure 1, the "Remove Hypothesized Adversity" block thus contained two mitigations: the shadow mitigation proposed in the second loop as well as the field-of-view mitigation proposed in the first.

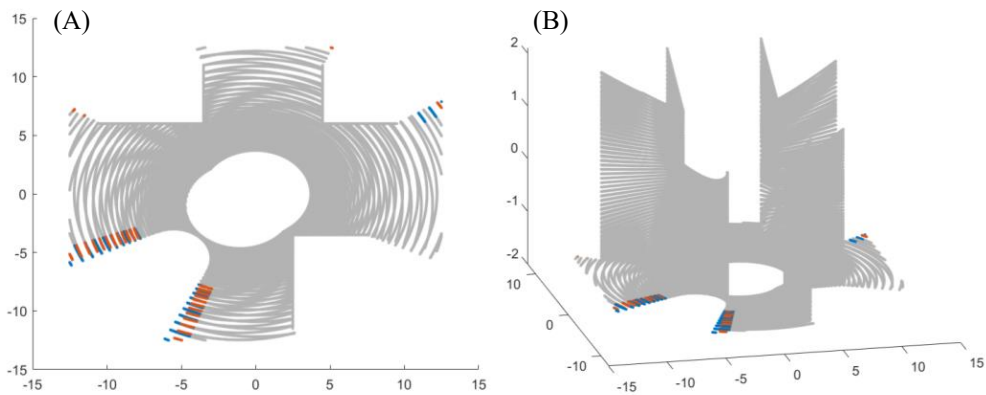

**FIGURE 10**
*Points in each cloud identified as lying in a shadow associated with the other cloud, shown (A) from above and (B) in perspective.*

This second mitigation step eliminated the discrepancy vectors associated with the cylinder, but not those associated with the building on the opposite corner (upper right of Figure 9A). The improvement is visualized in Figure 11, which shows the vector field after mitigation. The improvement is evident in the lower-left corner of Figure 11A, where the large vectors associated with the cylinder shadow have disappeared.

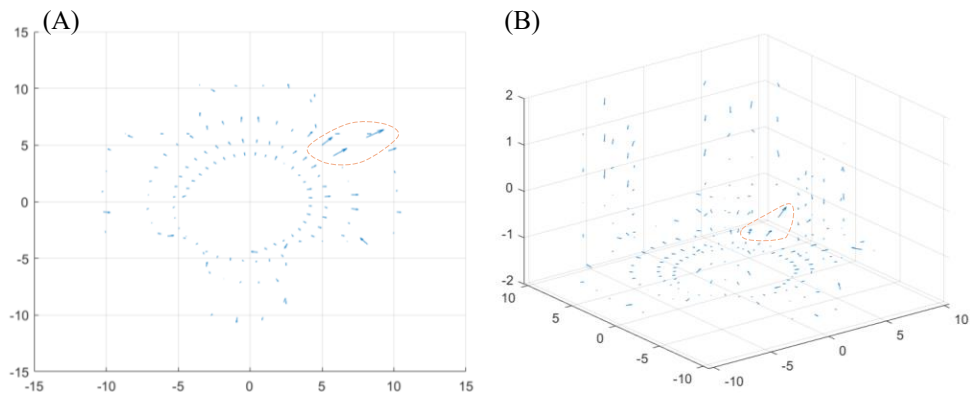

**FIGURE 11**
*Third discrepancy-vector field, after shadow removal.*

While discrepancy vectors remain in the vector field shown in Figure 11, a considerable impact has been made to reason about and mitigate adversities. In concept, the iterations could be continued, with the next goal of shrinking the large residual discrepancies identified in the upper-right corner of Figure 11
Figure 11A.

### 3.3 Experimental Data

To further evaluate the proposed analysis process, we applied the methodology to real-world data. In particular, we used USDOT-furnished data depicting a parking lot located near Fairfax, VA. The scene is shown in Figure 12. The data collection system (Wassaf et al., 2021) used a Velodyne VLP-16 lidar, with sixteen vertical channels spanning a field-of-view from -15° to +15° elevation. We considered one image pair from a larger dataset. This point-cloud pair were separated by an interval of 1 s, during which the vehicle moved in the y-direction at approximately 3 m/s. Due to limitations in GPS accuracy, the scans were aligned using NDT lidar scan matching (Biber & Straßer, 2003). The aligned scans are shown in Figure 13.

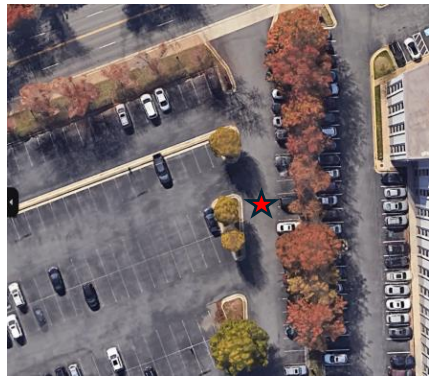

**FIGURE 12**
*Parking lot where experimental lidar data were acquired, with approximate vehicle location indicated by a red star.*

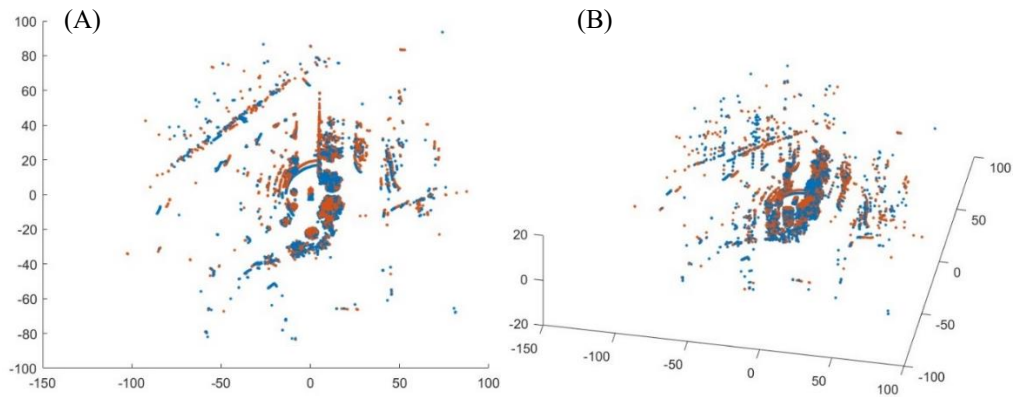

**FIGURE 13**
*Two experimental point clouds (red and blue), after registration, shown (A) from above and (B) in perspective.*

### 3.4 Application of Visualization Methodology to Experimental Data

The methodology of Figure 1 was again applied. Given the smaller number of vertical channels, a grid was chosen with 36 spherical voxels in the azimuth direction and 5 in the elevation direction. A discrepancy-vector field was computed across the two experimentally acquired point clouds. Figure 14 shows a visual representation of the initial discrepancy field.

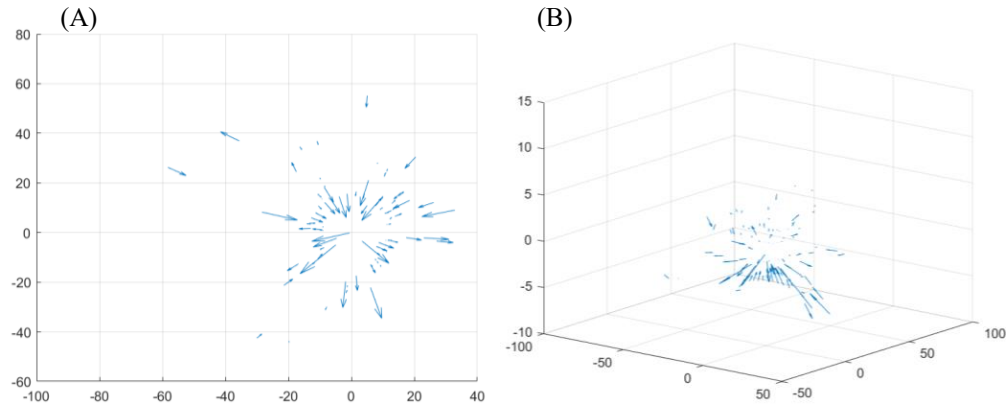

**FIGURE 14**
*Initial discrepancy-vector field generated from experimental data. The maximum discrepancy vector magnitude is 17.6 m.*

In the initial discrepancy field, the largest vectors are near the center of the point cloud, where lidar beams can reflect from the roof of the data collection van. Previous research shows that own-vehicle removal before the scan matching process increases localization accuracy (Jeong *et al.*, 2023). Own-vehicle removal was conducted by eliminating points within a radius of 3 m from the lidar center; the result was a cleaner discrepancy-vector field, as shown in Figure 15.

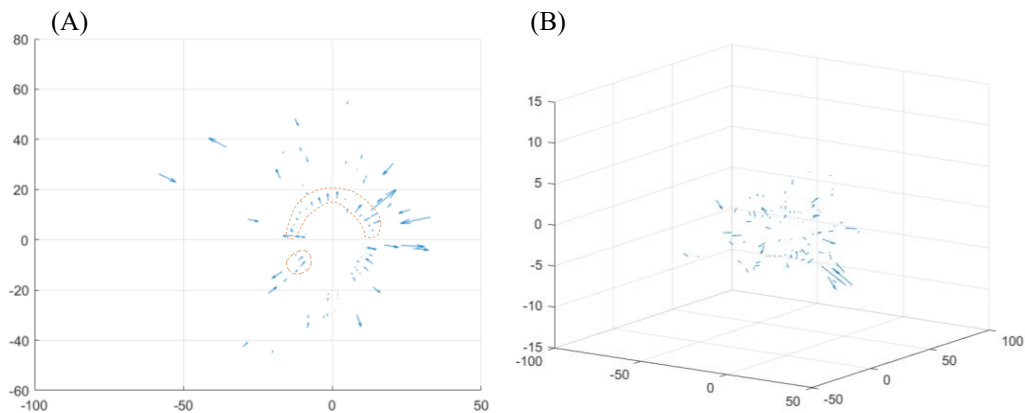

**FIGURE 15**
*Second discrepancy-vector field, with own-vehicle removed. The maximum discrepancy vector magnitude is 13.3 m.*

After own-vehicle removal, several large discrepancies persist near domain boundaries. These discrepancies form a familiar ring pattern around the field-of-view boundary. This ring is identified in Figure 15 with a red-dashed contour. As in the simulation-based analysis, points were removed at the domain boundaries for one scan if those points fell outside the lidar field-of-view for the location of the second scan. After domain-boundary mitigation, a third set of discrepancy vectors was computed, as shown in Figure 16A. The large discrepancy vectors have disappeared within the analyst-defined contours of Figure 15 (contours that are shown in green in Figure 16A).

In the third discrepancy-vector field, the analyst identified a pattern of large discrepancies on or around object contours, suggesting shadowing effects. Voxels seemingly associated with shadow effects are identified by red-dashed contours in Figure 16B. After applying the shadow-mitigation method of McDermott and Rife (2022), a fourth discrepancy-vector field resulted (see Figure 17).

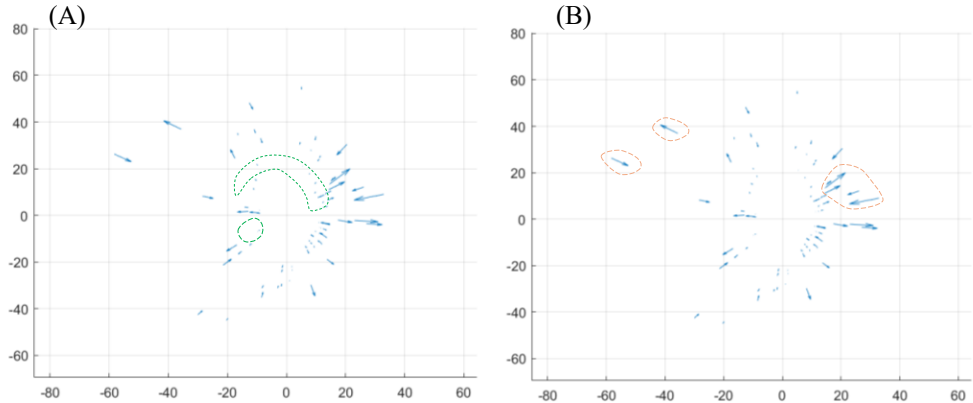

**FIGURE 16**
*Third discrepancy-vector field, after field-of-view mitigation. (A) The discrepancy vectors in the ring-shaped contours are eliminated by field-of-view mitigation. (B) A pattern of residual errors is evident near the edge of objects, like trees, as indicated by a new set of contours. The maximum discrepancy vector magnitude is 11.3 m.*

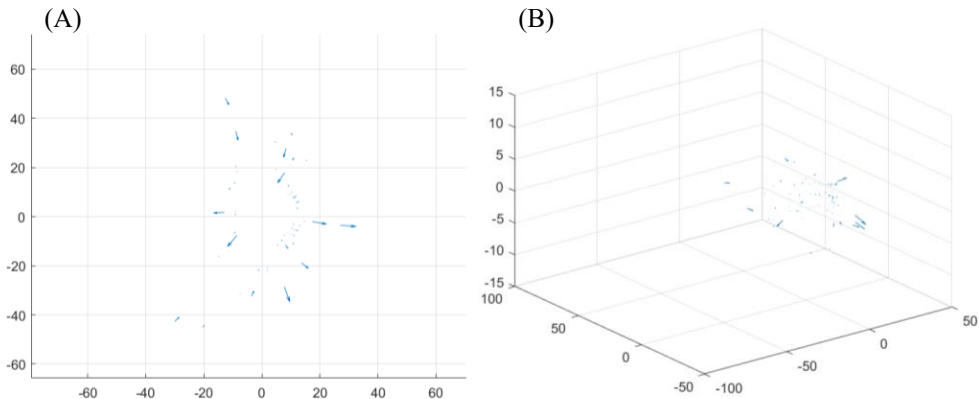

**FIGURE 17**
*Fourth discrepancy-vector field, after shadow mitigation. The maximum discrepancy vector magnitude is 8.1 m.*

While several lidar adversities have been addressed through this process, there remain a number of large discrepancies in Figure 17. Specifically, a set of long discrepancy vectors pointed radially inward or outward are visible. Another round of reasoning would be required on the part of the human analyst to identify and, ideally, mitigate this effect.

## 4  DISCUSSION

Developing reliable uncertainty quantification for lidar scan-matching algorithms remains an open question, in part because of the large number of systematic errors that appear during processing. Many of these adversities are related to the assumption that point clouds that depict the same static scene from different viewpoints are equivalent, given that they are correctly registered via a rigid transformation. It is not easy to spot these discrepancies when viewing raw point-cloud data, as seen in Figure 13; by contrast, the discrepancies become much more salient when viewed in the format of a vector field, as seen in Figure 14 through Figure 17. In this sense, our two examples provide a basic proof-of-concept for the utility of analyzing discrepancy-vector fields.

One salient point is that neither proof-of-concept example was iterated to completion. Ideally the final vector field would be small in magnitude and random in direction, as hypothesized in Figure 2; however, large, patterned discrepancies remained visible on the final iteration (third iteration for simulation, fourth iteration for experimental data). As context, we acknowledge the proof-of-concept examples demonstrated mitigation of known adversities, previously described in the research literature.

These adversities included ego-vehicle, field-of-view, and shadowing effects. Recognizing and characterizing new adversities is significantly more challenging, and so we leave that process to future work.

A curious aspect about implementing the algorithm is that the process appears to be relatively insensitive to the order in which adversities are identified and mitigated. In general, we expect that the human analyst will tend to focus on the largest visible discrepancies. However, there may not be a clear adversity mode that explains all of the largest discrepancy vectors. As seen from the experimental data in Figure 15A, for example, domain-boundary effects do not necessarily have larger magnitude discrepancies than the shadowing effects in Figure 16B. Fortunately, either of these issues can be addressed before the other without substantial impact on the form of the vector field once both are removed.

A final observation is that discrepancy patterns appear to become more complex as lidar resolution decreases. The simulation example considers a lidar sensor model with 80 horizontal beams across 32 degrees of vertical channels. By comparison, the experimental data utilizes a Velodyne VLP-16 lidar with 16 vertical beams across a similar elevation range. In effect, the elevation-direction resolution of the experimental data is roughly five times coarser than that of the simulation data. The relatively low angular resolution of the VLP-16 results in larger discrepancies in point patterns when imaging the same region of space from different vantage points. Though high-resolution may mitigate some adversities, low-resolution analysis may actually be beneficial for identification of lidar error sources, since adversities are more pronounced.

Additional future work may include the conversion of our error visualization strategy into a real-time failure diagnosis methodology. This would entail leveraging new rules and hypotheses from a human analyst into a direct pipeline for an online mitigation algorithm. The integrity of a real time mitigation system would improve by continuously adding new tools and deploying validated hypotheses within real-world datasets.

## 5 SUMMARY

In summary, we have presented an innovative technique for offline identification of lidar adversities that degrade scan matching, through a vector field visualization. The main contribution of this method includes the ability to troubleshoot and assist in localizing impactful discrepancies within a specific scene. Also, this method presents the ability to visualize representative adversities in lidar point clouds, to aid in their explanation. Our approach provides an interpretable visualization which differs from previous work in error mitigation strategies, which have generally proposed mitigations and then tested the effect on pose estimation, the final outcome of scan matching, rather than on the lidar scan itself, which is the input to scan matching.

## ACKNOWLEDGEMENTS

The authors wish to acknowledge and thank the U.S. Department of Transportation Joint Program Office (ITS JPO) and the Office of the Assistant Secretary for Research and Technology (OST-R) for sponsorship of this work. We also gratefully acknowledge SAIC and Tufts University, which supported specific aspects of this research. Opinions discussed here are those of the authors and do not necessarily represent those of the DOT, SAIC, or other affiliated agencies.